\documentclass[journal]{IEEEtran}
\usepackage{graphicx}
\usepackage{subfigure}
\usepackage{array}
\usepackage{amsmath}
\usepackage{amssymb}
\usepackage{multirow}
\usepackage{cases}
\usepackage{url}
\usepackage{cite}
\usepackage{fancyhdr}
\usepackage{mathrsfs}
\usepackage[justification=centering]{caption}
\usepackage[colorlinks=true,citecolor=black,linkcolor=black,anchorcolor=black,urlcolor=black,bookmarks=false]{hyperref}
\usepackage{epstopdf}

\ifCLASSINFOpdf

\else

\fi

\begin{document}

\title{Blind Stereoscopic Omnidirectional Image Quality Assessment Using Predictive Coding Hierarchy}

\author{Wei Zhou,~\IEEEmembership{Senior Member,~IEEE}, and André Kaup,~\IEEEmembership{Fellow, IEEE}
\thanks{W. Zhou is with the School of Computational and Mathematical Sciences, Cardiff University, CF24 4AG Cardiff, United Kingdom (e-mail: zhouw26@cardiff.ac.uk).

A. Kaup is with the Chair of Multimedia Communications and Signal Processing, Friedrich-Alexander-Universität Erlangen-Nürnberg, 91058 Erlangen, Germany (e-mail: andre.kaup@fau.de).}
}

\markboth{IEEE Transactions on Multimedia}
{Shell \MakeLowercase{\textit{et al.}}: Bare Demo of IEEEtran.cls for IEEE Journals}

\maketitle

\begin{abstract}
Stereoscopic omnidirectional images (SOIs) have provided users with newly immersive quality of experience in virtual reality environments. However, developing efficient and accurate perceptual quality assessment metrics for SOIs remains challenging due to many factors such as freely changeable field of views and binocular vision. In this paper, based on the characteristics of the human visual system (HVS), we propose a Predictive Coding Hierarchy-inspired metric (PCH) for blind/no-reference stereoscopic omnidirectional image quality assessment. Motivated by the viewing process of SOIs, the proposed PCH includes a local cyclopean perception module, a global predictive perception module, and a visual quality regressor. First, observers browse different spherical sceneries from viewports, and aggregate the local visual information to infer the perceptual quality of SOIs. Therefore, we extract various viewports, followed by cyclopean conversion and saliency detection to approach the perception and attention of the human brain. After the local aggregation, viewers then infer the global scene in their minds. Based on the binocular mechanism, we fuse left and right views to perform predictive coding hierarchy modelling. Finally, the visual quality regressor is exploited to obtain the ultimate quality score related to both local and global perceptual cues. Extensive experiments demonstrate that the proposed PCH achieves competitive and consistently improved performance compared with state-of-the-art quality assessment methods.
\end{abstract}


\IEEEpeerreviewmaketitle

\section{Introduction}

\IEEEPARstart{W}ITH the rapid development of metaverse and virtual reality (VR) technology, there has been a growing amount of VR content in recent years \cite{zhang2023qoe}. As one of the most popular forms among VR content, stereoscopic omnidirectional images (SOIs) create immersive 3D/stereoscopic plus 360-degree experience for end-users \cite{ribeiro2017quality,hou2020predictive}. However, during the acquisition, compression, transmission, and display of such visual content, various quality issues may arise, leading to the degradation in the perceived quality of experience (QoE) \cite{xu2020state}. To monitor and optimize the perceptual QoE of consumers, it is urgently important to measure the visual quality of SOIs.

\begin{figure}[t]
	\centering
    \captionsetup{justification=justified}
    \includegraphics[width=9cm]{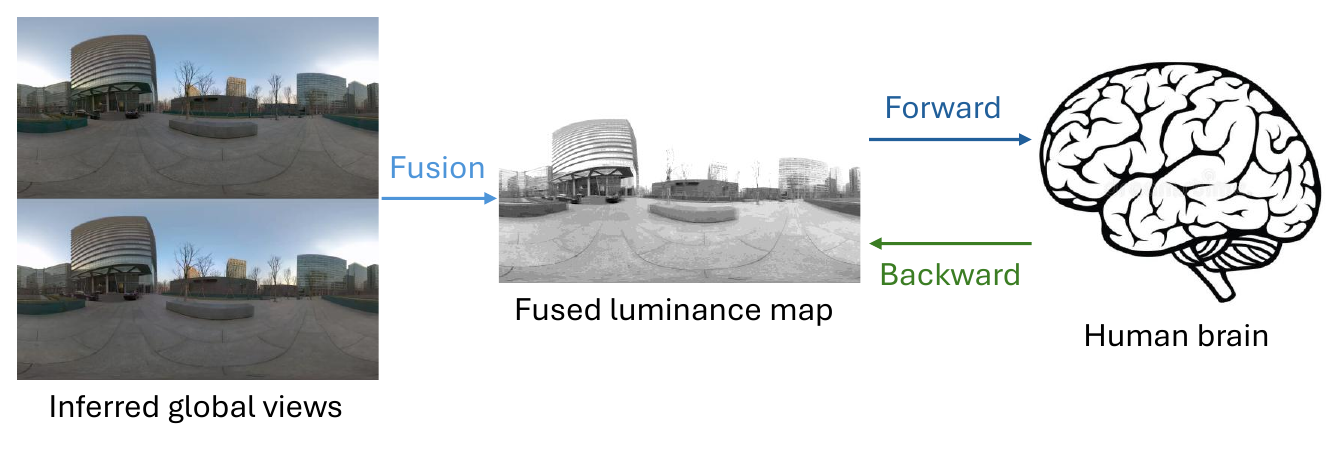}
	\caption{Illustration of perceiving a stereoscopic omnidirectional image with predictive coding hierarchy. We adopt luminance fusion since distortions are more sensitive in luminance.}
	\label{fig1}
\end{figure}

For image quality assessment (IQA) methods, the most accurate way is to design and conduct subjective tests \cite{wu2013crowdsourcing}. The collected subjective ratings together with images usually serve as the benchmark databases. However, subjective quality evaluation is time-consuming, expensive, and laborious, although crowdsourcing and ranking-based strategies have been explored to improve the efficiency and robustness of subjective assessment \cite{xu2012online,xu2013online}. Objective quality assessment offers an effective alternative, which automatically predicts the perceptual quality of images and thus can be easily integrated into multimedia systems for real-time quality monitoring and optimization \cite{liu2022liqa,jia2024bit}. In literature, objective quality assessment methods generally are divided into three categories, including full-reference (FR), reduced-reference (RR), and no-reference (NR). The FR IQA metrics rely on full information of the corresponding original/reference images, such as fidelity-related methods. The RR IQA approaches only exploit part of the reference, i.e., side information, while the NR IQA infers the perceptual quality without any reference features. In practical real-world applications, reference information is often unavailable or difficult to access, making NR IQA highly desirable for deployment \cite{sun2022graphiqa}.

Beyond the traditional plane images, each SOI is represented by a pair of omnidirectional images with left and right views, integrating the characteristics of both stereoscopic images and omnidirectional images. Compared with conventional planar IQA, blind SOI quality assessment is more challenging due to several unique factors \cite{zhou2024perceptual}. First, omnidirectional content is typically browsed through viewports with freely changeable fields of view (FoVs), where users only perceive a local region at each moment rather than the entire spherical scene. Second, binocular vision mechanisms, such as binocular fusion, rivalry, and disparity-related perception, further influence the perceived quality of SOIs. Third, observers tend to integrate local perceptual evidence and form a global mental representation of the scene, which implies that both local and global perceptual cues contribute to the final quality judgment.

Motivated by the above viewing process, we illustrate the perception of SOIs using predictive coding hierarchy in Fig. \ref{fig1}. predictive coding theory provides a biologically plausible explanation of hierarchical perception, where the brain continuously generates top-down predictions and refines them using bottom-up sensory evidence \cite{rao1999predictive,friston2005theory}. For SOIs, observers first browse different spherical sceneries through viewports and aggregate local visual information, and then infer a coherent global scene representation in their minds. In particular, we fuse the left and right views to construct a fused luminance map, since the human visual system (HVS) is more sensitive to luminance-based structural degradations than chrominance variations \cite{wang2004image,chen2017blind}. In predictive coding, feedforward pathways convey prediction errors, while feedback pathways provide updated predictions to iteratively refine perception \cite{bastos2012canonical}. Such a hierarchical inference mechanism suggests that perception emerges from the interaction between multi-level representations. Inspired by this perspective, we model SOI quality by integrating local viewport-based perceptual evidence with a global scene-level representation.

In this paper, according to the characteristics of the HVS, we propose a Predictive Coding Hierarchy-inspired metric (PCH) for blind/no-reference stereoscopic omnidirectional image quality assessment. Specifically, the proposed PCH consists of a local cyclopean perception module, a global predictive perception module, and a visual quality regressor. In the local cyclopean perception module, multiple viewports are extracted to simulate the browsing process, followed by cyclopean conversion and saliency detection to approximate binocular fusion and attention mechanisms. After local aggregation, the global predictive perception module models holistic scene-level cues by fusing left and right views to construct a fused luminance map and performing predictive coding hierarchy modelling. Finally, the visual quality regressor integrates both local and global perceptual features to predict the ultimate quality score.

The main contributions of this work are summarized as follows:
\begin{itemize}
\item We propose a Predictive Coding Hierarchy-inspired blind/no-reference metric for stereoscopic omnidirectional image quality assessment, which explicitly models both local viewport perception and global scene perception in a unified framework.
\item We design a local cyclopean perception module based on viewport extraction, cyclopean conversion, and saliency-aware statistical modelling, enabling effective characterization of attention-driven perceptual quality cues in the local cyclopean domain.
\item We introduce a global predictive perception module based on binocular luminance fusion and predictive coding hierarchy modelling, and experiments demonstrate that our proposed method consistently achieves competitive and improved performance compared with representative state-of-the-art quality assessment methods.
\end{itemize}

The rest of this paper is organized as follows. Section II reviews related work. Section III presents the proposed PCH framework. Section IV reports experimental results and analysis. Finally, Section V concludes this paper.

\section{Related Work}
In this section, we briefly review the relevant studies on omnidirectional image quality assessment, stereoscopic image quality assessment, blind/no-reference IQA, as well as predictive-coding-inspired perceptual modelling.

\subsection{Omnidirectional Image Quality Assessment}
As virtual reality applications are increasingly popular, omnidirectional images have attracted increasing research attention in recent years. Different from conventional planar images, omnidirectional images are typically represented by spherical projections such as equirectangular projection (ERP), which introduces non-uniform sampling density and projection-induced distortions \cite{zhou2021no}. Therefore, directly applying traditional 2D IQA methods to omnidirectional images often leads to unsatisfactory performance.

Existing omnidirectional IQA methods can generally be categorized into viewport-based approaches and spherical-domain approaches. Viewport-based methods \cite{xu2020blind,jiang2021multi} extract multiple viewports from omnidirectional images according to specific sampling strategies and then aggregate local quality cues, which is consistent with the fact that users perceive omnidirectional content through multiple FoVs. Spherical-domain methods \cite{chen2018spherical,duan2018perceptual} attempt to directly model distortions on the sphere by considering geometric properties and projection characteristics.

Although significant progress has been achieved, most existing omnidirectional IQA methods focus on monoscopic omnidirectional images and do not explicitly account for binocular perception mechanisms in stereoscopic omnidirectional images.

\subsection{Stereoscopic Image Quality Assessment}
Stereoscopic image quality assessment has been widely studied due to its importance in 3D multimedia applications \cite{wang2015quality,liu2017binocular}. Compared with 2D images, stereoscopic images involve binocular vision mechanisms such as binocular fusion, rivalry, disparity perception, and depth-related discomfort \cite{yang2020latitude}. 

A common strategy in stereoscopic IQA is to construct cyclopean representations that approximate the fused perceptual view of the HVS \cite{chen2013full}. In addition, disparity and binocular rivalry features have also been incorporated to better characterize stereoscopic perceptual quality \cite{zhou2019dual,xu2020binocular}.

However, stereoscopic IQA methods designed for planar images cannot be directly applied to SOIs, since SOIs exhibit additional challenges including omnidirectional browsing behavior, spherical distortions, and global scene integration across viewports.

\subsection{Blind/No-Reference IQA and HVS-Inspired Modelling}
Blind/no-reference IQA has received extensive attention because reference information is often unavailable in practical applications. Early NR IQA approaches were mainly based on natural scene statistics (NSS), where distortions are quantified by measuring deviations from statistical regularities of natural images \cite{mittal2012making,mittal2012no}. In addition, model-based blind IQA methods have also been explored by explicitly characterizing distortion-specific perceptual responses, such as the dual-model framework for noisy image quality assessment \cite{zhai2015dual}.

More recently, learning-based NR IQA methods have achieved promising performance by exploiting deep neural networks \cite{ma2017end}, perceptual feature learning \cite{zhang2018blind}, deep meta-learning \cite{zhu2020metaiqa}, distortion graph representations \cite{sun2022graphiqa}, and opinion-unaware learning from multiple annotators \cite{wang2025deep}. In addition, active fine-tuning from gMAD examples has been shown to improve the generalization ability of blind IQA models by identifying challenging counterexamples and refining model behavior accordingly \cite{wang2021active}. Meanwhile, HVS-inspired modelling has also been explored in NR IQA methods, such as visual attention mechanisms, contrast sensitivity, and hierarchical processing \cite{liu2019pre,zhai2013retina}. 

For omnidirectional and stereoscopic content, HVS characteristics become even more important due to viewport-based perception and binocular mechanisms. Nevertheless, existing blind SOI quality assessment methods still lack an explicit modelling of the perceptual process that integrates local browsing evidence into a global scene representation.

\begin{figure*}[t]
	\centering
    \captionsetup{justification=justified}
    \includegraphics[width=16.5cm]{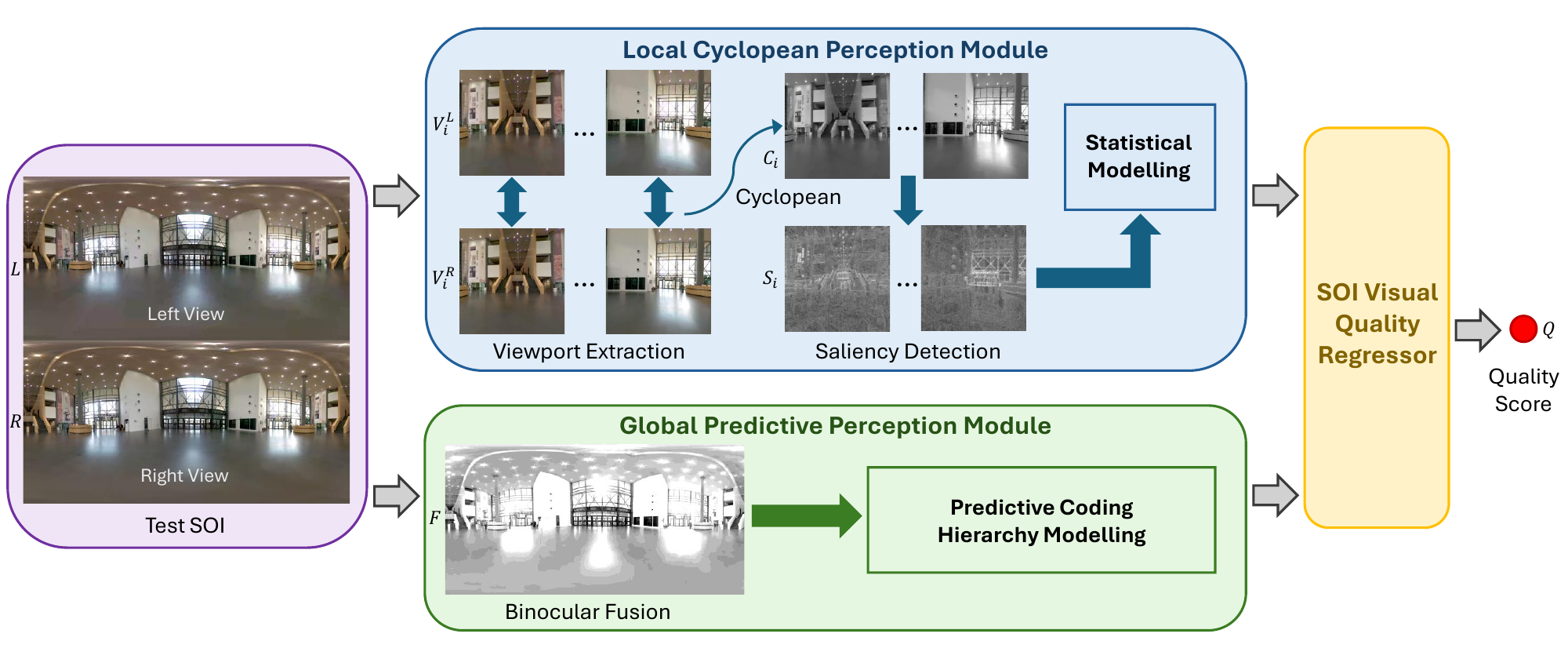}
	\caption{The pipeline of our proposed PCH for blind stereoscopic omnidirectional image quality assessment.}
	\label{fig2}
\end{figure*}

\subsection{Predictive Coding Theory and Perceptual Hierarchy}
Predictive coding theory provides a biologically plausible explanation of perception as hierarchical inference \cite{huang2011predictive,clark2013whatever}. In predictive coding, higher-level cortical areas generate top-down predictions of sensory inputs, while lower-level areas compute prediction errors between actual inputs and predicted signals. These prediction errors are propagated forward, and updated predictions are fed back iteratively to minimize the overall mismatch.

Such a forward-backward hierarchical mechanism has been widely used to interpret visual perception, efficient coding, and neural information processing \cite{spratling2017review}. In the context of image quality assessment, predictive coding representations have also been explored to capture distortion-induced irregularities through error responses and coefficient statistics \cite{zhai2013comparative,chen2020stereoscopic}.

Motivated by the hierarchical perceptual mechanism of predictive coding, we propose to model SOI quality perception through both local viewport perception and global predictive coding hierarchy modelling. This allows us to better approximate the human viewing process of stereoscopic omnidirectional images. In addition, to the best of our knowledge, the proposed method is one of the first blind/no-reference metrics considering predictive coding theory and perceptual hierarchy for SOI quality assessment.

Although SOIQE \cite{xu2020binocular} also adopts predictive coding theory for stereoscopic omnidirectional image quality assessment, it is fundamentally different from the proposed framework in several aspects. First, SOIQE is a full-reference metric relying on both pristine and distorted images, whereas the proposed PCH is a blind/no-reference metric without access to any reference information. Second, SOIQE mainly models predictive coding responses between reference and distorted stimuli, while PCH performs hierarchical perceptual modelling directly on distorted stereoscopic omnidirectional content through global predictive inference. Third, the proposed method additionally incorporates cyclopean perception modelling, saliency-aware attention statistics and binocular luminance fusion, enabling unified local-global perceptual quality modelling.

\section{Proposed Method}
In this section, we introduce the proposed Predictive Coding Hierarchy-inspired framework (PCH) for blind/no-reference stereoscopic omnidirectional image quality assessment. 
As illustrated in Fig.~\ref{fig2}, PCH is motivated by the hierarchical perceptual process of human observers when viewing stereoscopic omnidirectional images (SOIs). 
Specifically, observers first browse local viewports with attention allocation, and then integrate binocular information into a global scene representation through hierarchical predictive inference. Accordingly, the proposed PCH consists of three main components, including a local cyclopean perception module, a global predictive perception module, and a visual quality regressor.

Given a stereoscopic omnidirectional image pair $(L, R)$, the goal of PCH is to predict its perceptual quality score $Q$ without access to any reference information.

\subsection{Overview of the Proposed Framework}
The overall pipeline of PCH can be summarized as follows. First, multiple viewport pairs are extracted from the distorted stereoscopic omnidirectional image to simulate the browsing behavior of users. Then, cyclopean perception modelling and saliency detection are applied to characterize local perceptual quality cues. After local aggregation, a global fused luminance representation is constructed from binocular views, and predictive coding hierarchy modelling is performed to capture holistic perceptual regularities and distortion-induced abnormalities. Finally, both local and global descriptors are integrated and mapped to the final quality score via a regression model. The local cyclopean perception module mainly captures viewport-level perceptual degradations, saliency-related attention variations, and local binocular inconsistencies within currently attended FoVs. In contrast, the global predictive perception module models scene-level structural irregularities and holistic binocular perceptual inference. Since human observers perceive stereoscopic omnidirectional content hierarchically by progressively integrating local perceptual evidence into a global scene representation, the two modules provide complementary perceptual quality characterization.

\subsection{Local Cyclopean Perception Module}
The local cyclopean perception module aims to simulate the viewport-based perception process of SOIs \cite{wan2025no}. Unlike conventional planar images, omnidirectional images are perceived through multiple FoVs, where observers sequentially explore different viewports and allocate attention to salient regions.

\subsubsection{Viewport Extraction}
Given the stereoscopic omnidirectional image pair $(L, R)$, we extract $N$ viewport pairs:
\begin{equation}
\{(V_i^L, V_i^R)\}_{i=1}^{N},
\end{equation}
\noindent where each $(V_i^L, V_i^R)$ corresponds to the left and right viewport images sampled from different viewing directions \cite{chen2020stereoscopic}. In this work, we adopt $N=6$ viewports as the default setting, which provides an effective trade-off between prediction accuracy and computational efficiency. Increasing the number of viewports further only yields marginal performance improvement while increasing computational complexity, as further validated in Section IV-E.

\subsubsection{Cyclopean Perception Modelling}
In stereoscopic omnidirectional image perception, the HVS does not process the left and right views independently.
Instead, binocular fusion integrates the two slightly different retinal inputs into a unified perceptual representation, commonly referred to as the cyclopean view.
Such cyclopean perception plays an essential role in stereoscopic quality judgment, especially under binocular rivalry and asymmetric distortions.

Given the $i$-th stereoscopic viewport pair extracted from the left and right omnidirectional images, denoted as $V_i^L$ and $V_i^R$, we aim to construct a perceptually consistent cyclopean viewport $C_i$ that approximates binocular fusion. Since accurate fusion relies on reliable inter-view correspondence, a disparity map is first estimated between the two viewports.

For each pixel location $(x,y)$, the disparity $d_i(x,y)$ is computed using an SSIM-based stereo matching strategy \cite{chen2013full}, which selects the displacement that maximizes local structural similarity between the left and right views.
Using the estimated disparity map, the right viewport is warped into the coordinate system of the left viewport. Specifically, the warped coordinate is defined as
\begin{equation}
y' = y - d_i(x,y).
\end{equation}

The disparity-compensated right view $\tilde{V}_i^R$ is obtained via linear interpolation:
\begin{equation}
\begin{aligned}
\tilde{V}_i^R(x,y) =\;&
\big(y'-\lfloor y'\rfloor\big)\,
V_i^R\!\left(x,\lfloor y'\rfloor+1\right) \\
&+\big(\lfloor y'\rfloor+1-y'\big)\,
V_i^R\!\left(x,\lfloor y'\rfloor\right),
\end{aligned}
\end{equation}

Psychophysical studies indicate that regions with stronger spatial activity tend to dominate binocular perception, particularly under rivalry conditions. To characterize local spatial reliability, an information response map is computed for each viewport. For a grayscale image $V$, the local mean and variance are estimated using Gaussian smoothing:
\begin{equation}
\mu(x,y)=G_\sigma \ast V(x,y), \quad
\sigma^2(x,y)=G_\sigma \ast V(x,y)^2 - \mu(x,y)^2,
\end{equation}
where $G_\sigma$ denotes a Gaussian kernel.
The information response is then defined as
\begin{equation}
I(x,y)=\log_2\!\left(1+\sigma^2(x,y)\right).
\end{equation}

Accordingly, the information response maps $I_i^L$ and $I_i^R$ are computed from $V_i^L$ and $V_i^R$, respectively. The information response of the right view is aligned using the same disparity compensation:
\begin{equation}
\tilde{I}_i^R(x,y)=I_i^R(x,y-d_i(x,y)).
\end{equation}

Finally, the cyclopean viewport is synthesized using information-guided fusion. We define the normalization term as
\begin{equation}
Z_i(x,y)= I_i^L(x,y)+\tilde{I}_i^R(x,y).
\end{equation}
The fusion weights are computed as
\begin{equation}
w_L(x,y)=\frac{I_i^L(x,y)}{Z_i(x,y)},
\end{equation}
\begin{equation}
w_R(x,y)=\frac{\tilde{I}_i^R(x,y)}{Z_i(x,y)}.
\end{equation}

The resulting cyclopean representation is obtained by
\begin{equation}
C_i(x,y)=w_L(x,y)\,V_i^L(x,y)+w_R(x,y)\,\tilde{V}_i^R(x,y).
\end{equation}

To ensure numerical stability, zero-valued information responses are replaced by a small constant. Through the above disparity-aware and information-guided fusion process, the constructed cyclopean viewport provides an effective approximation of local binocular perception and serves as the basis for subsequent saliency detection and local feature extraction.

\subsubsection{Saliency Detection and Attention Modelling}
In addition to the cyclopean perception modelling, observers tend to focus on visually distinctive regions, while distortions in non-salient areas contribute less to the perceptual quality. Therefore, saliency detection is incorporated to approximate visual attention mechanisms.

Specifically, we first obtain a downsampled representation of the cyclopean viewport $C_i$ using low-pass filtering and uniform subsampling:
\begin{equation}
\tilde{C}_i(p,q)=A \ast C_i^{s}(pr,qr),
\end{equation}
where $(p,q)$ denotes pixel coordinates in the downsampled domain. $A$ represents a low-pass filter, and $\ast$ is the convolution operation.


The downsampled viewport is then transformed into the frequency domain using the discrete cosine transform (DCT):
\begin{equation}
F_i=\mathrm{DCT}(\tilde{C}_i).
\end{equation}
Following the image signature-based saliency formulation \cite{hou2011image,zhou2023reduced}, only the sign of the DCT coefficients is preserved:
\begin{equation}
\hat{F}_i=\mathrm{sign}(F_i).
\end{equation}

Finally, the saliency map is obtained by applying the inverse DCT and squaring the response:
\begin{equation}
S_i=\left(\mathrm{IDCT}(\hat{F}_i)\right)^2.
\end{equation}
The resulting saliency map highlights visually distinctive regions and provides an effective attention prior for local quality perception.

\subsubsection{Local Feature Extraction and Aggregation}
After cyclopean fusion and saliency detection, we obtain a saliency map $S_i$ for each stereoscopic viewport pair. The saliency map characterizes the spatial distribution of visual attention and highlights perceptually dominant regions. Instead of extracting features directly from the cyclopean intensity, we compute lightweight statistical descriptors from the saliency map itself inspired by \cite{poreddy2021no}. This design aims to characterize the overall attention distribution pattern, which is closely related to perceptual quality degradation.

Given a saliency map $S_i \in \mathbb{R}^{h \times w}$, three statistics are extracted:
\begin{equation}
f_i^{(1)} = \mu(S_i),
\qquad
f_i^{(2)} = \sigma(S_i),
\qquad
f_i^{(3)} = \mathcal{H}(S_i),
\end{equation}
where $\mu(\cdot)$ and $\sigma(\cdot)$ denote the mean and standard deviation over all pixels, respectively, and $\mathcal{H}(\cdot)$ denotes the Shannon entropy computed on 8-bit quantized saliency values for stable histogram estimation. Specifically, the mean reflects the overall saliency strength, the standard deviation characterizes the spatial dispersion of perceptual attention, and the entropy measures the structural complexity of salient regions. Distortions often alter the spatial distribution and organization of visually salient regions, making these statistics effective for perceptual quality characterization. The resulting local feature vector is defined as
\begin{equation}
\mathbf{f}_i^{\mathrm{local}} =
\left[
f_i^{(1)},\,
f_i^{(2)},\,
f_i^{(3)}
\right]^{\top}.
\end{equation}

The extracted saliency-based statistics reflect the attention strength, dispersion, and complexity of perceptual responses. Distortions often alter the spatial distribution of salient regions, leading to measurable changes in these statistics. When multiple viewports are used, we aggregate the local descriptors by concatenating statistics extracted from all saliency maps.

\subsection{Global Predictive Perception Module}
While the local cyclopean perception module focuses on viewport-level binocular fusion and attention, human observers further integrate local evidence and form a coherent global scene representation. Such a process is consistent with predictive coding hierarchy, where perception emerges from iterative interactions between feedforward prediction errors and feedback updated predictions. Accordingly, we construct a global perceptual stimulus from the stereoscopic pair and extract predictive coding-based features to characterize distortion-induced global irregularities.

\subsubsection{Binocular Luminance Fusion}
We first construct a global fused luminance map from the stereoscopic omnidirectional image pair. Given the left and right omnidirectional views $L$ and $R$, we extract their luminance components:
\begin{equation}
Y^L = \mathcal{G}(L), \qquad
Y^R = \mathcal{G}(R),
\end{equation}
where $\mathcal{G}(\cdot)$ denotes grayscale conversion.
Binocular fusion is then performed by absolute summation:
\begin{equation}
F = \left|Y^L + Y^R\right|.
\end{equation}
The fused luminance map enhances binocular-consistent structures and highlights distortion-sensitive variations, forming a compact global perceptual stimulus.

\subsubsection{Multi-scale Lateral Geniculate Nucleus-Inspired Preprocessing}
Inspired by the multi-scale receptive fields in the lateral geniculate nucleus (LGN) \cite{wandell1995foundations}, we adopt a set of downsampling factors to model multi-resolution visual perception:
\begin{equation}
\mathcal{S}=\{s_k\}_{k=1}^{4}, \qquad s_k=2^{k-1}.
\end{equation}
In our experiments, four scales are used corresponding to
$s_k\in\{1,2,4,8\}$. Such multi-scale modelling is motivated by the hierarchical receptive field properties of the HVS, where visual perception operates across different spatial resolutions. For each scale $s_k$, we obtain a downsampled luminance map by uniform subsampling:
\begin{equation}
F^{(s_k)}(p,q)=F(s_k p,s_k q).
\end{equation}

To mimic the center-surround response and nonlinear activation observed in the early visual system \cite{croner1995receptive}, each scaled map is filtered by a Laplacian-of-Gaussian (LoG) kernel followed by a hyperbolic tangent nonlinearity:
\begin{equation}
X^{(s_k)}=\tanh\!\left(F^{(s_k)} * h_{\mathrm{LoG}}\right),
\end{equation}
where $*$ denotes convolution and $h_{\mathrm{LoG}}$ is a LoG filter.
This step enhances structural cues while suppressing extreme responses.

\subsubsection{Block-wise Predictive Coding Inference}
While the predictive coding formulation is related to the full-reference metric \cite{chen2020stereoscopic}, the proposed framework differs in both input representation and objective. Specifically, \cite{chen2020stereoscopic} employs predictive coding to model neural response properties under a full-reference setting, whereas the proposed method performs block-wise predictive coding inference on multi-scale stereoscopic omnidirectional luminance maps after binocular luminance fusion and LGN-inspired preprocessing, aiming to extract distortion-sensitive latent coefficients and residual responses without requiring reference images.

For each scale $s_k$, the preprocessed response map $X^{(s_k)}$ is partitioned into non-overlapping blocks of size $b\times b$, 
where $b=16$ as set in \cite{chen2020stereoscopic}. Each block is vectorized to form
\begin{equation}
\mathbf{X}^{(s_k)}=
\left[
\mathbf{x}^{(s_k)}_1,
\dots,
\mathbf{x}^{(s_k)}_{B_k}
\right]
\in\mathbb{R}^{m\times B_k},
\end{equation}
where $m=b^2$ and $B_k$ denotes the number of blocks at scale $s_k$. Let $\mathbf{U}\in\mathbb{R}^{m\times K}$ be a fixed dictionary learned offline.
Predictive coding explains the sensory input using latent coefficients
$\mathbf{R}^{(s_k)}\in\mathbb{R}^{K\times B_k}$.
The top-down prediction is defined as
\begin{equation}
\hat{\mathbf{X}}^{(s_k)}
=
\tanh\!\left(
\mathbf{U}\mathbf{R}^{(s_k)}
\right),
\end{equation}
and the feedforward prediction error is
\begin{equation}
\mathbf{E}^{(s_k)}
=
\mathbf{X}^{(s_k)}
-
\hat{\mathbf{X}}^{(s_k)}.
\end{equation}

Following predictive coding theory, the coefficients are iteratively updated to minimize prediction errors. At iteration $t$, the feedback refinement is expressed as
\begin{equation}
\begin{aligned}
\mathbf{R}^{(s_k)} \leftarrow\;&
\mathbf{R}^{(s_k)}
+
\frac{k_1}{\sigma}
\mathbf{U}^{\top}
\Big(
\mathbf{G}^{(s_k)}
\odot
\mathbf{E}^{(s_k)}
\Big) \\
&-
k_1 \alpha
\frac{
\mathbf{R}^{(s_k)}
}{
1+\mathbf{R}^{(s_k)}\odot\mathbf{R}^{(s_k)}
},
\end{aligned}
\end{equation}
where $\odot$ denotes element-wise multiplication,
$k_1$, $\sigma$, and $\alpha$ are fixed parameters,
and the gating term
\begin{equation}
\mathbf{G}^{(s_k)}
=
1-
\hat{\mathbf{X}}^{(s_k)}
\odot
\hat{\mathbf{X}}^{(s_k)}
\end{equation}
corresponds to the derivative of the hyperbolic tangent nonlinearity. After several iterations, the optimized coding coefficients
$\mathbf{R}^{(s_k)}$ and residual errors
$\mathbf{E}^{(s_k)}$ are obtained,
which jointly characterize top-down predictions and bottom-up irregularities.

\subsubsection{Global Feature Extraction}
Different from the full-reference metric \cite{chen2020stereoscopic}, which focuses on neural response modeling between reference and distorted images, we further aggregate the block-wise coding coefficients and residual errors across spatial blocks and scales to construct global descriptors for perceptual quality prediction. That is, for each scale $s_k$, we summarize predictive coding outputs by averaging across blocks:
\begin{equation}
\mathbf{v}^{(s_k)}=
\frac{1}{B_k}\sum_{j=1}^{B_k}
\mathbf{R}^{(s_k)}_{:,j},
\qquad
\mathbf{e}^{(s_k)}=
\frac{1}{B_k}\sum_{j=1}^{B_k}
\mathbf{E}^{(s_k)}_{:,j}.
\end{equation}

Finally, multi-scale global features are formed by concatenation:
\begin{equation}
\begin{aligned}
\mathbf{v} &=
\left[
\mathbf{v}^{(s_1)},
\mathbf{v}^{(s_2)},
\mathbf{v}^{(s_3)},
\mathbf{v}^{(s_4)}
\right], \\
\mathbf{e} &=
\left[
\mathbf{e}^{(s_1)},
\mathbf{e}^{(s_2)},
\mathbf{e}^{(s_3)},
\mathbf{e}^{(s_4)}
\right].
\end{aligned}
\end{equation}

The resulting global representation
$\left[\mathbf{v},\mathbf{e}\right]$ jointly captures top-down coding responses and bottom-up residual irregularities, which are correlated with perceptual quality degradation in stereoscopic omnidirectional images.

\subsection{Visual Quality Regressor}
After extracting local cyclopean perception features and global predictive-coding representations, we predict the final perceptual quality score via a supervised regression model.

\subsubsection{Feature Fusion}
For each stereoscopic omnidirectional image, the proposed framework produces two complementary feature sets, including local features (i.e., $\mathbf{f}_{\mathrm{local}}=[\mathbf{f}_i^{\mathrm{local}}]$) extracted from cyclopean viewports with saliency-aware statistical descriptors and global features (i.e., $\mathbf{f}_{\mathrm{global}}=[\mathbf{v},\mathbf{e}]$) obtained from multi-scale predictive coding inference.

The final feature vector is constructed by concatenation:
\begin{equation}
\mathbf{f}=
\big[
\mathbf{f}_{\mathrm{local}},
\mathbf{f}_{\mathrm{global}}
\big].
\end{equation}
In our implementation, the global representation consists of predictive coding coefficients and prediction errors averaged across blocks at four scales, while the local representation consists of statistics (mean, standard deviation, and entropy) computed from the saliency maps.

\begin{table}[t]
\centering
\captionsetup{justification=justified}
\caption{Performance comparison with existing methods on SOLID database, where IQA includes methods for both planar and stereoscopic content.}
\label{tab:comparison}
\renewcommand{\arraystretch}{1.2}
\setlength{\tabcolsep}{6pt}
\begin{tabular}{l|l|c|c|c}
\hline
\textbf{Category} & \textbf{Methods} & \textbf{SRCC} $\uparrow$ & \textbf{PLCC} $\uparrow$ & \textbf{RMSE} $\downarrow$ \\
\hline

\multirow{5}{*}{FR IQA} 
& PSNR   & 0.847 & 0.862 & 0.399 \\
& SSIM \cite{wang2004image}   & 0.866 & 0.850 & 0.408 \\
& FSIM \cite{zhang2011fsim}   & 0.883 & 0.845 & 0.420 \\
& GMSD \cite{xue2013gradient}   & 0.899 & 0.894 & 0.346 \\
& MJ3DQA \cite{chen2013full}  & 0.877 & 0.847 & 0.410 \\
\hline

\multirow{4}{*}{FR OIQA} 
& S-PSNR \cite{yu2015framework}   & 0.856 & 0.870 & 0.389 \\
& WS-PSNR \cite{sun2017weighted}  & 0.851 & 0.817 & 0.435 \\
& CPP-PSNR \cite{zakharchenko2016quality} & 0.853 & 0.870 & 0.388 \\
& SOIQE \cite{chen2020stereoscopic}    & 0.914 & 0.914 & \textbf{0.318} \\
\hline

\multirow{11}{*}{NR IQA} 
& NIQE \cite{mittal2012making}     & 0.734 & 0.760 & 0.509 \\
& HOSA \cite{xu2016blind}     & 0.633 & 0.726 & 0.544 \\
& dipIQ \cite{ma2017dipiq}    & 0.611 & 0.620 & 0.603 \\
& BPRI \cite{min2017blind}     & 0.629 & 0.654 & 0.596 \\
& BRISQUE \cite{mittal2012no}  & 0.796 & 0.802 & 0.464 \\
& OG \cite{liu2016blind}       & 0.701 & 0.694 & 0.555 \\
& MEON \cite{ma2017end}     & 0.799 & 0.780 & 0.467 \\
& DB-CNN \cite{zhang2018blind}   & 0.866 & 0.845 & 0.416 \\
& MetaIQA \cite{zhu2020metaiqa}  & 0.799 & 0.774 & 0.490 \\
& UNIQUE \cite{zhang2021uncertainty}     & 0.467 & 0.525 & 0.659 \\
& SINQ \cite{liu2017binocular}       & 0.779 & 0.810 & 0.460 \\
\hline

\multirow{4}{*}{NR OIQA} 
& VP-BSOIQA \cite{qi2020viewport}  & 0.842 & 0.853 & 0.411 \\
& MBIDCN \cite{chai2021monocular}     & 0.872 & 0.879 & 0.372 \\
& BPGI \cite{liu2025bpgi}       & 0.929 & 0.931 & 0.366 \\
& \textbf{Proposed PCH} & \textbf{0.938} & \textbf{0.947} & \textbf{0.318} \\
\hline

\end{tabular}
\end{table}

\begin{table*}[t]
\centering
\captionsetup{justification=justified}
\caption{Performance comparison under different distortion settings, including symmetric/asymmetric distortions and JPEG/BPG compression schemes. Note that IQA includes methods for both planar and stereoscopic content.}
\label{tab:overall_comparison}
\renewcommand{\arraystretch}{1.2}
\setlength{\tabcolsep}{6pt}

\resizebox{\textwidth}{!}{
\begin{tabular}{l|l|cc|cc|cc|cc}
\hline
\multirow{2}{*}{Category} & \multirow{2}{*}{Methods}
& \multicolumn{2}{c|}{Symmetric Distortion}
& \multicolumn{2}{c|}{Asymmetric Distortion}
& \multicolumn{2}{c|}{JPEG Compression}
& \multicolumn{2}{c}{BPG Compression} \\
\cline{3-10}
& & SRCC$\uparrow$ & PLCC$\uparrow$
  & SRCC$\uparrow$ & PLCC$\uparrow$
  & SRCC$\uparrow$ & PLCC$\uparrow$
  & SRCC$\uparrow$ & PLCC$\uparrow$ \\
\hline

\multirow{5}{*}{FR IQA}
& PSNR   & 0.807 & 0.902 & 0.747 & 0.779 & 0.831 & 0.861 & 0.881 & 0.898 \\
& SSIM \cite{wang2004image}   & 0.823 & 0.889 & 0.778 & 0.823 & 0.898 & 0.921 & 0.901 & 0.920 \\
& FSIM \cite{zhang2011fsim}   & 0.837 & 0.914 & 0.807 & 0.840 & 0.910 & 0.919 & 0.911 & 0.931 \\
& GMSD \cite{xue2013gradient}   & 0.851 & 0.927 & 0.830 & 0.844 & 0.908 & 0.937 & 0.917 & 0.931 \\
& MJ3DQA \cite{chen2013full} & 0.803 & 0.940 & 0.783 & 0.822 & 0.870 & 0.912 & 0.875 & 0.925 \\
\hline

\multirow{4}{*}{FR OIQA}
& S-PSNR \cite{yu2015framework}  & 0.816 & 0.849 & 0.760 & 0.807 & 0.846 & 0.843 & 0.882 & 0.860 \\
& WS-PSNR \cite{sun2017weighted}  & 0.815 & 0.905 & 0.744 & 0.795 & 0.841 & 0.817 & 0.883 & 0.868 \\
& CPP-PSNR \cite{zakharchenko2016quality} & 0.815 & 0.909 & 0.749 & 0.797 & 0.842 & 0.780 & 0.884 & 0.925 \\
& SOIQE \cite{chen2020stereoscopic}    & 0.868 & 0.948 & 0.846 & 0.853 & 0.905 & 0.922 & 0.919 & 0.902 \\
\hline

\multirow{11}{*}{NR IQA}
& NIQE \cite{mittal2012making}     & 0.756 & 0.858 & 0.473 & 0.592 & 0.813 & 0.860 & 0.855 & 0.860 \\
& HOSA \cite{xu2016blind}     & 0.638 & 0.765 & 0.368 & 0.508 & 0.810 & 0.878 & 0.653 & 0.747 \\
& dipIQ \cite{ma2017dipiq}    & 0.663 & 0.713 & 0.478 & 0.521 & 0.739 & 0.737 & 0.605 & 0.611 \\
& BPRI \cite{min2017blind}     & 0.751 & 0.811 & 0.289 & 0.318 & 0.800 & 0.766 & 0.690 & 0.738 \\
& BRISQUE \cite{mittal2012no}  & 0.820 & 0.924 & 0.636 & 0.620 & 0.820 & 0.840 & 0.868 & 0.869 \\
& OG \cite{liu2016blind}       & 0.753 & 0.855 & 0.526 & 0.507 & 0.687 & 0.677 & 0.800 & 0.813 \\
& MEON \cite{ma2017end}     & 0.803 & 0.803 & 0.775 & 0.736 & 0.816 & 0.789 & 0.790 & 0.800 \\
& DB-CNN \cite{zhang2018blind}   & 0.860 & 0.930 & 0.776 & 0.768 & 0.902 & 0.908 & 0.881 & 0.893 \\
& MetaIQA \cite{zhu2020metaiqa}  & 0.827 & 0.846 & 0.640 & 0.598 & 0.877 & 0.887 & 0.866 & 0.858 \\
& UNIQUE \cite{zhang2021uncertainty}     & 0.390 & 0.565 & 0.426 & 0.488 & 0.424 & 0.458 & 0.567 & 0.669 \\
& SINQ \cite{liu2017binocular}       & 0.822 & 0.926 & 0.598 & 0.660 & 0.812 & 0.843 & 0.813 & 0.826 \\
\hline

\multirow{4}{*}{NR OIQA}
& VP-BSOIQA \cite{qi2020viewport}  & 0.834 & 0.909 & 0.698 & 0.717 & 0.907 & 0.930 & 0.839 & 0.844 \\
& MBIDCN \cite{chai2021monocular}     & 0.818 & 0.933 & 0.779 & 0.776 & 0.883 & 0.921 & 0.882 & 0.892 \\
& BPGI \cite{liu2025bpgi}       & \textbf{0.934} & \textbf{0.972} & 0.887 & 0.898 & \textbf{0.937} & 0.948 & 0.931 & 0.940 \\
& \textbf{Proposed PCH} & 0.906 & \textbf{0.972}
& \textbf{0.902} & \textbf{0.914}
& \textbf{0.937} & \textbf{0.955}
& \textbf{0.940} & \textbf{0.961} \\
\hline
\end{tabular}
}
\end{table*}

\subsubsection{SVR-based Quality Prediction}
Given the fused feature vector $\mathbf{f}$, we employ an $\varepsilon$-support vector regression (SVR) model with an RBF kernel to map features to subjective quality scores:
\begin{equation}
Q=\mathcal{R}(\mathbf{f}),
\end{equation}
where $\mathcal{R}(\cdot)$ denotes the learned regression function and $Q$ is the predicted quality score.

The SVR with RBF kernel is adopted due to its strong nonlinear fitting capability and robustness under relatively limited training data, which is common in current IQA databases. This regressor effectively integrates local perceptual evidence and global predictive coding irregularities, yielding accurate blind stereoscopic omnidirectional image quality prediction.

\section{Experimental Results and Analysis}
In this section, we first introduce the experimental settings, followed by comprehensive comparisons with state-of-the-art methods. Further analyses are conducted under different distortion conditions and compression schemes to demonstrate the robustness of PCH. In addition, ablation studies are provided to validate the effectiveness of each component.

\subsection{Experimental Settings}
Following the evaluation protocol in \cite{chen2020stereoscopic}, experiments are conducted on three publicly available quality assessment databases to comprehensively evaluate the proposed PCH.

Among them, SOLID is adopted as the primary benchmark for stereoscopic omnidirectional IQA, since it specifically contains distorted omnidirectional stereoscopic images. In addition, two conventional stereoscopic IQA databases, namely LIVE Phase I and Phase II, are included to evaluate the robustness of the proposed predictive coding hierarchy modelling across different stereoscopic content modalities.

\subsubsection{SOLID \cite{xu2018subjective}}
The primary evaluation is performed on a stereoscopic omnidirectional image quality assessment database called SOLID, which contains 276 distorted SOIs with JPEG or BPG compression. There exist 84 symmetrically and 192 asymmetrically distorted images in this database. Each stereoscopic omnidirectional image is associated with Mean Opinion Score (MOS) values collected via standardized subjective testing procedures.

\subsubsection{LIVE Phase I \cite{moorthy2013subjective}}
This database contains 20 pristine stereoscopic image pairs and 365 symmetrically distorted stereoscopic images. Five common distortion types are involved in the database, including JPEG compression, JPEG2000 compression, Gaussian blur, white noise, and fast fading. Subjective scores are provided in the form of differential mean opinion scores (DMOS).

\subsubsection{LIVE Phase II \cite{chen2013full}}
LIVE Phase II extends Phase I with additional distortion configurations and asymmetric distortion scenarios between the left and right views. The database is composed of 8 original stereoscopic images and 360 distorted stereoscopic image pairs. In total, there exist 120 symmetrically distorted and 240 asymmetrically distorted stereoscopic images. Each distorted image also corresponds to a DMOS value.

For objective evaluation, we follow common practices in IQA and report the Spearman rank-order correlation coefficient (SRCC), Pearson linear correlation coefficient (PLCC), and root mean square error (RMSE). Higher SRCC and PLCC values, as well as lower RMSE values, indicate better performance. Note that PLCC and RMSE are computed after a nonlinear regression as recommended in the VQEG protocol. For each database, we adopt an 80\%-20\% random train-test split strategy for 100 times and report the median SRCC, PLCC, and RMSE values.

\begin{table}[t]
\centering
\captionsetup{justification=justified}
\caption{Ablation study on module contribution.}
\label{tab:ablation_module}
\renewcommand{\arraystretch}{1.3}
\setlength{\tabcolsep}{6pt}
\begin{tabular}{l|c|c|c}
\hline
Methods & SRCC$\uparrow$ & PLCC$\uparrow$ & RMSE$\downarrow$ \\
\hline
Local Cyclopean Perception Module & 0.917 & 0.922 & 0.381 \\
Global Predictive Perception Module & 0.908 & 0.919 & 0.402 \\
\textbf{Proposed PCH}              & \textbf{0.938} & \textbf{0.947} & \textbf{0.318} \\
\hline
\end{tabular}
\end{table}

\begin{table}[t]
\centering
\captionsetup{justification=justified}
\caption{Influence of viewport number and saliency modelling for the proposed method.}
\label{tab:viewport_saliency}
\renewcommand{\arraystretch}{1.3}
\setlength{\tabcolsep}{8pt}
\begin{tabular}{l|c|c|c}
\hline
Methods & SRCC$\uparrow$ & PLCC$\uparrow$ & RMSE$\downarrow$ \\
\hline
6 viewports (without saliency)  & 0.911 & 0.924 & 0.380 \\
Proposed PCH (6 viewports)      & 0.938 & 0.947 & 0.318 \\
20 viewports (without saliency) & 0.922 & 0.929 & 0.374 \\
Proposed PCH (20 viewports)     & \textbf{0.940} & \textbf{0.949} & \textbf{0.310} \\
\hline
\end{tabular}
\end{table}

\begin{table}[t]
\centering
\captionsetup{justification=justified}
\caption{Performance comparison on LIVE Phase I.}
\label{tab:live_phase1_comparison}
\renewcommand{\arraystretch}{1.3}
\setlength{\tabcolsep}{10pt}
\begin{tabular}{l|c|c|c}
\hline
\textbf{Methods} & \textbf{SRCC}$\uparrow$ & \textbf{PLCC}$\uparrow$ & \textbf{RMSE}$\downarrow$ \\
\hline
You \cite{you2010perceptual}        & 0.814 & 0.830 & 7.746 \\
Benoit \cite{benoit2009quality}     & 0.878 & 0.881 & 7.061 \\
Hewage \cite{hewage2010reduced}     & 0.899 & 0.902 & 9.139 \\
MJ3DQA \cite{chen2013full}   & 0.916 & 0.917 & 6.533 \\
Chen \cite{chen2013no}   & 0.891 & 0.895 & 7.247 \\
Bensalma \cite{bensalma2013perceptual}   & 0.875 & 0.887 & 7.559 \\
SOIQE \cite{chen2020stereoscopic}      & 0.917 & 0.920 & 6.266 \\
\textbf{Proposed PCH} & \textbf{0.918} & \textbf{0.934} & \textbf{5.901} \\
\hline
\end{tabular}
\end{table}

\begin{table}[t]
\centering
\captionsetup{justification=justified}
\caption{Performance comparison on LIVE Phase II.}
\label{tab:live_phase2_comparison}
\renewcommand{\arraystretch}{1.3}
\setlength{\tabcolsep}{10pt}
\begin{tabular}{l|c|c|c}
\hline
\textbf{Methods} & \textbf{SRCC}$\uparrow$ & \textbf{PLCC}$\uparrow$ & \textbf{RMSE}$\downarrow$ \\
\hline
You \cite{you2010perceptual}        & 0.786 & 0.800 & 6.772 \\
Benoit \cite{benoit2009quality}     & 0.728 & 0.748 & 7.490 \\
Hewage \cite{hewage2010reduced}     & 0.501 & 0.558 & 9.364 \\
MJ3DQA \cite{chen2013full}   & 0.889 & 0.900 & 4.987 \\
Chen \cite{chen2013no}   & 0.880 & 0.895 & 5.102 \\
Bensalma \cite{bensalma2013perceptual}   & 0.751 & 0.770 & 7.204 \\
SOIQE \cite{chen2020stereoscopic}      & 0.907 & 0.915 & 4.544 \\
\textbf{Proposed PCH} & \textbf{0.913} & \textbf{0.922} & \textbf{4.405} \\
\hline
\end{tabular}
\end{table}

\subsection{Overall Performance Comparison}
Table~\ref{tab:comparison} reports the overall performance comparison between the proposed PCH and existing full-reference (FR) and no-reference (NR) IQA and OIQA methods. The compared methods include traditional FR IQA metrics, omnidirectional FR metrics, learning-based NR IQA approaches, as well as recent NR OIQA models.

It can be observed that the proposed PCH achieves robust performance across all the evaluation settings. In particular, the proposed PCH achieves the highest SRCC of 0.938 and PLCC of 0.947, outperforming the strongest competing NR OIQA model BPGI. At the same time, its RMSE is highly competitive and comparable to the best-performing baselines. Moreover, the proposed PCH outperforms existing NR OIQA methods and even surpasses FR metrics, demonstrating its strong capability in predicting perceptual quality without reference information. These results validate the effectiveness of integrating local viewport perception and global predictive coding hierarchy modelling. Although the proposed global fusion strategy adopts a relatively simple binocular luminance integration scheme, the subsequent predictive coding inference effectively captures distortion-induced irregularities and binocular imbalance through residual responses. Meanwhile, the local cyclopean perception module explicitly incorporates disparity-aware fusion, which further improves robustness against asymmetric distortions.

\subsection{Performance Under Different Distortion Conditions}
To further investigate robustness under diverse degradation scenarios, we evaluate the compared methods under multiple distortion settings, including symmetric/asymmetric distortions and JPEG/BPG compression schemes. Symmetric distortions refer to the case where identical distortions are applied to both the left and right views, whereas asymmetric distortions correspond to situations where the two views are affected by different distortion types or distortion levels. The results are summarized in Table~\ref{tab:overall_comparison}.

From the table, we can see that the proposed PCH achieves consistently superior correlations across most evaluation scenarios. In particular, our PCH demonstrates strong robustness against binocular quality imbalance and asymmetric degradation between the left and right views. Additionally, under both JPEG and BPG compression, PCH maintains the best prediction accuracy, indicating its effectiveness in handling different compression-induced artifacts. These results confirm that integrating local cyclopean perception and global predictive coding hierarchy modelling provides a reliable and generalizable solution for blind stereoscopic omnidirectional image quality assessment.

\subsection{Ablation Study on Local and Global Modules}
To verify the contribution of each component in the proposed PCH framework, we conduct ablation experiments by evaluating the local cyclopean perception module and global predictive perception module separately. The results are reported in Table~\ref{tab:ablation_module}.

The local module alone achieves SRCC = 0.917, indicating that saliency-aware viewport perception provides strong quality cues. Meanwhile, the global predictive coding module also yields competitive performance (SRCC = 0.908), demonstrating the effectiveness of modelling holistic irregularities.

By combining both modules, our proposed PCH improves prediction accuracy, confirming that local attention-driven cues and global predictive coding representations are complementary. The performance gain of more than 2\% in SRCC over the local-only variant indicates that predictive coding modelling captures distortion-induced global irregularities that cannot be explained by saliency-based local statistics alone.

\subsection{Influence of Viewport Number and Saliency Modelling}
We further investigate the influence of viewport number and the importance of saliency modelling. Table~\ref{tab:viewport_saliency} shows results under different viewport configurations with and without saliency features.

The results demonstrate that removing saliency consistently degrades performance, verifying that attention modelling is essential for local quality perception. Increasing the number of viewports from 6 to 20 provides marginal improvement, but at the cost of higher computational complexity. Therefore, we adopt 6 viewports as the default setting for a favorable trade-off between efficiency and accuracy. In addition, since the proposed PCH is a non-deep-learning framework, reporting the number of trainable parameters is not applicable. Its computational cost mainly arises from viewport extraction, cyclopean perception modelling, saliency computation, and predictive coding inference, rather than network parameterization.

\subsection{Evaluation on LIVE Stereoscopic IQA Databases}
Following \cite{chen2020stereoscopic}, to further validate the generalization capability of the proposed PCH framework on conventional stereoscopic IQA scenarios, we conduct experiments on the LIVE stereoscopic IQA databases, including Phase I and Phase II. Since the LIVE databases contain conventional stereoscopic image pairs rather than omnidirectional content and viewport extraction is not applicable to stereoscopic images, the local cyclopean perception module is adapted accordingly. Specifically, instead of extracting spherical viewports, we employ patch-based sampling to simulate local perceptual processing. Furthermore, disparity information between the left and right views is estimated and incorporated to enhance global perceptual representation, enabling more accurate modeling of binocular interaction and depth-related distortions.

Tables~\ref{tab:live_phase1_comparison} and~\ref{tab:live_phase2_comparison}
report the overall performance comparisons. It can be observed that the proposed PCH consistently achieves the best correlation with subjective scores across both databases, outperforming existing stereoscopic IQA baselines as well as the representative SOI metric SOIQE.

We further investigate the influence of patch size in predictive coding modelling. Tables~\ref{tab:patch_size_phase1} and~\ref{tab:patch_size_phase2}
evaluate different patch configurations. From these tables, we observe that using medium-scale patches (i.e., a patch size of 128×128) yields the best rank and linear correlations (SRCC/PLCC) on both databases, while a smaller patch size can slightly reduce RMSE on LIVE Phase I. Considering overall correlation accuracy and stability, we adopt 128×128 as the default patch size in our proposed PCH method.

\begin{table}[t]
\centering
\captionsetup{justification=justified}
\caption{Influence of patch size in predictive coding modelling on LIVE Phase I.}
\label{tab:patch_size_phase1}
\renewcommand{\arraystretch}{1.3}
\setlength{\tabcolsep}{10pt}
\begin{tabular}{l|c|c|c}
\hline
\textbf{Patch Sizes} & \textbf{SRCC}$\uparrow$ & \textbf{PLCC}$\uparrow$ & \textbf{RMSE}$\downarrow$ \\
\hline
64 × 64  & 0.913 & 0.933 & \textbf{5.842} \\
128 × 128 & \textbf{0.918} & \textbf{0.934} & 5.901 \\
256 × 256 & 0.905 & 0.923 & 6.313 \\
\hline
\end{tabular}
\end{table}

\subsection{Discussion}
The comprehensive experimental results consistently validate the effectiveness
of the proposed predictive coding hierarchy framework. Unlike conventional viewport-based or feature-driven NR metrics, PCH explicitly models the perceptual inference process from local attention allocation to global predictive refinement.

\begin{table}[t]
\centering
\captionsetup{justification=justified}
\caption{Influence of patch size in predictive coding modelling on LIVE Phase II.}
\label{tab:patch_size_phase2}
\renewcommand{\arraystretch}{1.3}
\setlength{\tabcolsep}{10pt}
\begin{tabular}{l|c|c|c}
\hline
\textbf{Patch Sizes} & \textbf{SRCC}$\uparrow$ & \textbf{PLCC}$\uparrow$ & \textbf{RMSE}$\downarrow$ \\
\hline
64 × 64  & 0.896 & 0.910 & 4.656 \\
128 × 128 & \textbf{0.913} & \textbf{0.922} & \textbf{4.405} \\
256 × 256 & 0.907 & 0.920 & 4.425 \\
\hline
\end{tabular}
\end{table}

The ablation studies demonstrate that saliency-aware local statistics capture distortion-sensitive attention distribution, while predictive coding modelling characterizes global structural irregularities through error-driven inference. The complementary nature of these two mechanisms explains the performance gain when both modules are integrated.

Furthermore, experimental results on LIVE stereoscopic IQA databases verify that the proposed framework generalizes well beyond SOI scenarios, indicating that predictive coding hierarchy provides a robust and biologically plausible foundation for immersive visual quality modelling. These findings suggest that explicitly modelling hierarchical perceptual inference is a promising direction for blind quality assessment of complex immersive visual content.

\section{Conclusion}
In this paper, we have proposed a Predictive Coding Hierarchy-inspired blind quality assessment metric (PCH) for stereoscopic omnidirectional images. Motivated by the hierarchical perceptual mechanism of the human visual system, the proposed framework explicitly integrates both local viewport-based perception and global scene-level predictive inference. Specifically, a local cyclopean perception module is developed to simulate binocular fusion and attention allocation through cyclopean modelling and saliency-aware statistical descriptors. Meanwhile, a global predictive perception module is introduced to characterize holistic distortion-induced irregularities by binocular luminance fusion and multi-scale predictive coding representation. Finally, local and global perceptual cues were jointly fused and mapped to subjective quality scores.

Extensive experiments conducted on public quality assessment databases demonstrate that the proposed PCH achieves competitive and consistently improved performance compared with representative state-of-the-art methods. Further ablation studies validate the effectiveness and complementarity of each component, confirming that predictive-coding-inspired perceptual hierarchy modelling provides a reliable and generalizable solution for immersive visual quality assessment. Such a capability is valuable for perceptual QoE monitoring, content optimization, and quality-aware delivery in emerging virtual reality and immersive multimedia systems.

In the future, we plan to explore more efficient hierarchical feature learning strategies and extend the proposed framework toward video-based omnidirectional quality assessment scenarios, where temporal perceptual dynamics also play an important role \cite{barkowsky2009temporal}. Furthermore, incorporating depth perception, visual discomfort, and temporal attention mechanisms may further improve the modelling of overall quality of experience in virtual reality environments. Adaptive viewport sampling according to distortion characteristics or visual attention may also further improve the efficiency and robustness of the proposed framework and will be investigated in future work.



%





\ifCLASSOPTIONcaptionsoff
  \newpage
\fi



%
\bibliographystyle{IEEEtran}
\bibliography{tmmRefer}

\section*{Biography}
\vspace{-1cm}
\begin{IEEEbiography}
[{\includegraphics[width=1in,height=1.25in,clip,keepaspectratio]{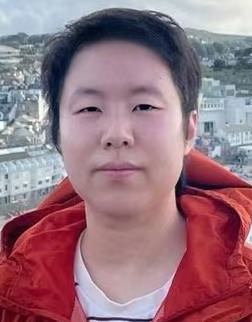}}]{Wei Zhou (S’19–M’21–SM’24)}
is an Associate Professor at Cardiff University, United Kingdom. Dr Zhou was a Postdoctoral Fellow at University of Waterloo, Canada. Wei received the Ph.D. degree from the University of Science and Technology of China in 2021, joint with the University of Waterloo from 2019 to 2021. Wei was a visiting professor at Dalian University of Technology, visiting scholar at National Institute of Informatics, Japan, a research assistant with Intel, and a research intern at Microsoft Research and Alibaba Cloud.

Dr Zhou is now an Associate Editor of IEEE Transactions on Neural Networks and Learning Systems, Pattern Recognition, and ACM Transactions on Multimedia Computing, Communications, and Applications. Wei’s research interests span multimedia computing, perceptual image processing, and computational vision.
\end{IEEEbiography}


\begin{IEEEbiography}
[{\includegraphics[width=1in,height=1.25in,clip,keepaspectratio]{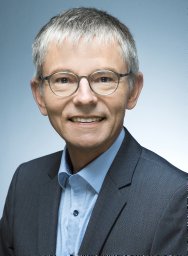}}]{André Kaup (Fellow, IEEE)} received the Dipl.-Ing. and Dr.-Ing. degrees in electrical engineering from RWTH Aachen University, Aachen, Germany, in 1989 and 1995, respectively.

He joined Siemens Corporate Technology, Munich, Germany, in 1995, and became the Head of the Mobile Applications and Services Group in 1999. Since 2001, he has been a Full Professor and the Head of the Chair of Multimedia Communications and Signal Processing at Friedrich-Alexander University Erlangen-Nürnberg (FAU), Germany. From 2005 to 2007 he was Vice Speaker of the DFG Collaborative Research Center 603. From 2015 to 2017, he served as the Head of the Department of Electrical Engineering and Vice Dean of the Faculty of Engineering at FAU. He has authored around 500 journal and conference papers and has over 120 patents granted or pending. His research interests include image and video signal processing and coding, and multimedia communication.

Dr. Kaup is vice-chair of the IEEE Image, Video, and Multidimensional Signal Processing Technical Committee and a member of the Scientific Advisory Board of the German VDE/ITG. He is an IEEE Fellow and a member of the Bavarian Academy of Sciences and Humanities, the German National Academy of Science and Engineering, and the European Academy of Sciences and Arts. He is a member of the Editorial Board of the IEEE Circuits and Systems Magazine. He was a Siemens Inventor of the Year 1998 and obtained the 1999 ITG Award. He received several IEEE best paper awards, including the Paul Dan Cristea Special Award in 2013, and his group won the Grand Video Compression Challenge from the Picture Coding Symposium 2013. The Faculty of Engineering with FAU and the State of Bavaria honored him with Teaching Awards, in 2015 and 2020, respectively. He served as an Associate Editor of the IEEE Transactions on Circuits and Systems for Video Technology. He was a Guest Editor of the IEEE Journal of Selected Topics in Signal Processing.
\end{IEEEbiography}



%

%
%
%




\end{document}